\documentclass[letterpaper, 10 pt, conference]{ieeeconf}  

\IEEEoverridecommandlockouts                              

\usepackage{cite}

\usepackage{graphics} 
\usepackage{epsfig} 
\usepackage{mathptmx} 
\usepackage{times} 
\usepackage{amsmath} 
\usepackage{amssymb}  
\usepackage{algorithm}
\usepackage{algpseudocode}
\usepackage{multirow} 
\usepackage{booktabs} 
\usepackage{float}          

\title{\LARGE \bf
Empart: Interactive Convex Decomposition for Converting Meshes to Parts
}

\author{Brandon Vu$^{1}$, Shameek Ganguly$^{1}$, Pushkar Joshi$^{1}$ 
\thanks{$^{1}$ Intrinsic Innovation LLC
        }
}

\begin{document}

\maketitle
\thispagestyle{empty}
\pagestyle{empty}

\newcommand{\ToolName}{Empart}

\begin{abstract}
Simplifying complex 3D meshes is a crucial step in robotics applications to enable efficient motion planning and physics simulation.
Common methods, such as approximate convex decomposition, represent a mesh as a collection of simple parts, which are computationally inexpensive to simulate. However, existing approaches apply a uniform error tolerance across the entire mesh, which can result in a sub-optimal trade-off between accuracy and performance. For instance, a robot grasping an object needs high-fidelity geometry in the vicinity of the contact surfaces but can tolerate a coarser simplification elsewhere. A uniform tolerance can lead to excessive detail in non-critical areas or insufficient detail where it's needed most.

To address this limitation, we introduce Empart, an interactive tool that allows users to specify different simplification tolerances for selected regions of a mesh. Our method leverages existing convex decomposition algorithms as a sub-routine but uses a novel, parallelized framework to handle region-specific constraints efficiently. Empart provides a user-friendly interface with visual feedback on approximation error and simulation performance, enabling designers to iteratively refine their decomposition. We demonstrate that our approach significantly reduces the number of convex parts compared to a state-of-the-art method (V-HACD) at a fixed error threshold, leading to substantial speedups in simulation performance. For a robotic pick-and-place task, Empart-generated collision meshes reduced the overall simulation time by 69\% compared to a uniform decomposition, highlighting the value of interactive, region-specific simplification for performant robotics applications.
\end{abstract}

\section{Introduction}
\label{sec:introduction}

\begin{figure}
    \centering
  
    \includegraphics[width=\columnwidth]{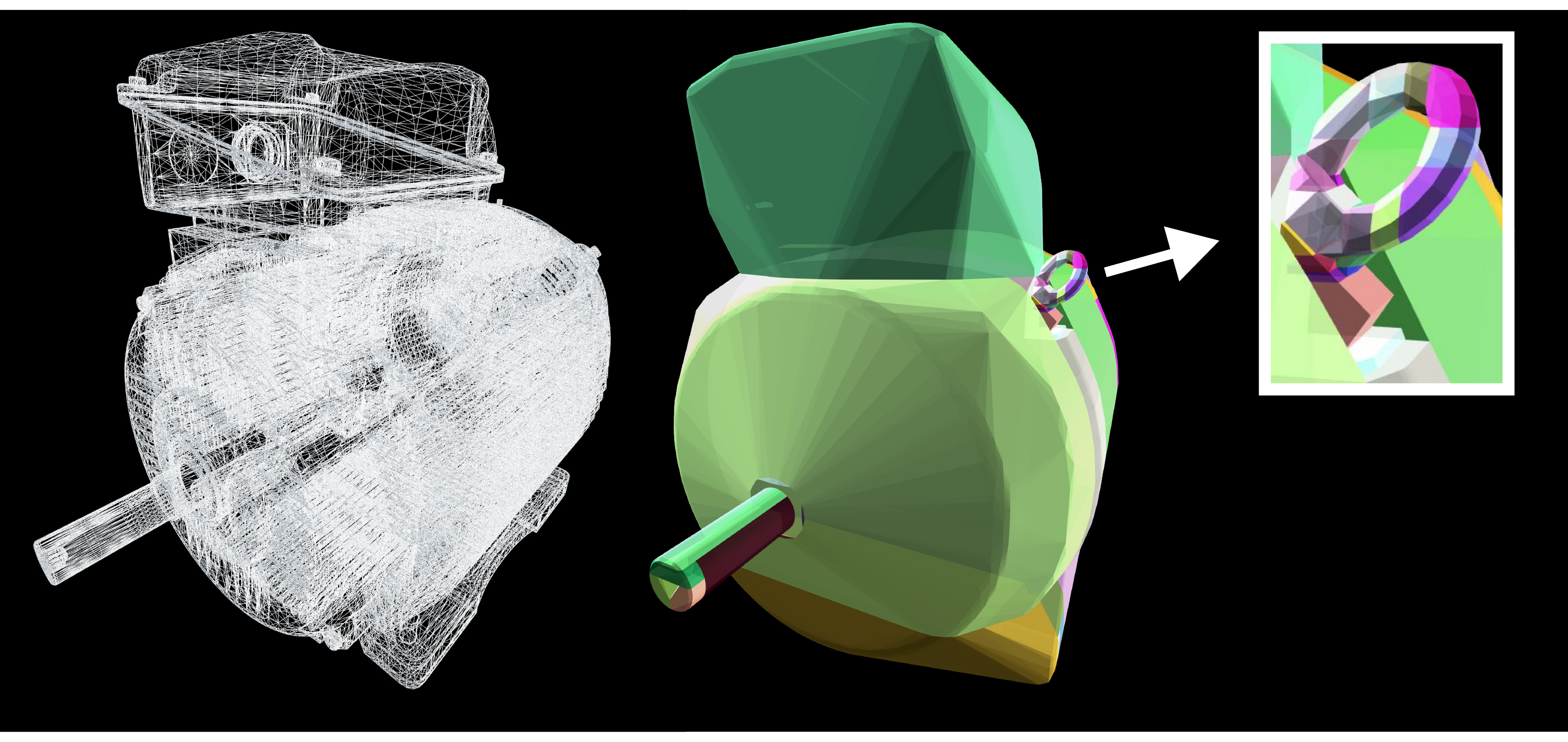}
  
    \caption{We introduce $\ToolName$, an interactive tool for simplifying 3D meshes via convex decomposition. $\ToolName$ allows users to specify error tolerances across different regions of the mesh, thereby allowing fine-grained control over decomposition. This example shows a dense mesh of an electric motor (left) and the output from $\ToolName$ with 166 convex parts (middle). User-specified regions of the object, such as the lifting eye (right), are preserved with higher detail.
} 
    \label{fig:teaser}
\end{figure}

Software for robotics, video games, and computer-generated movies often requires digital models of physical objects. A common task is to simulate the movement and interaction of physical objects with each other and with their environment. For this simulation to be realistic, we need to compute accurate collisions between models. Accurate collision detection is computationally expensive for models of non-trivial shape, which makes it prohibitively slow for many practical applications. Simplification of the shape allows for faster collision detection and is therefore a necessary step for the useful simulation of real-world objects.   

Shapes are often represented as meshes, i.e. a collection of polygons embedded in 3D space. Shape simplification then boils down to mesh simplification, a process which accepts a mesh as input and produces another mesh as output. The output mesh often has far fewer polygons than the input mesh, which makes it faster to compute accurate collisions with other shapes. The output mesh may be composed of sub-meshes, each a very simple shape primitive. Convex decomposition, that is, representing the input mesh as a collection of convex pieces, is a common form of mesh simplification. Without loss of generality, we consider only mesh simplification via convex decomposition in this paper. 

In order to be useful for collision detection, the simplified mesh needs to be a sufficiently faithful representation of the input shape. Here, ``sufficiently faithful representation'' is task-specific. For example, if the task is to conservatively avoid an object, the simplified mesh may simply be a convex hull of the input mesh. However, if the task is to simulate the manipulation of the input shape with multiple contact points (e.g., a robot arm grasping and inserting a connector into a socket), the simplified mesh must have sufficient detail in the contact regions. 

Depending on the task, the same input mesh may need to be simplified differently. In fact, a subset of the input mesh may need to be preserved as-is, with simplification being performed for the rest. For this reason, it is highly valuable to allow the user to control the mesh simplification such that necessary details are preserved while unnecessary ones are removed. The $\ToolName$ tool described in this paper allows such a workflow.

\textbf{Contributions} In this paper we present a novel algorithm for convex decomposition that allows interactive mesh simplification. The simplified mesh consists of significantly fewer convex parts compared to the state-of-the-art in convex decomposition, for the same level of user-specified detail. As a result, we observe significant speedups in the applications that require collision detection, such as robot simulation. Our algorithm is made accessible through an open-source implementation, that includes a web-based mesh simplification tool.

\section{Related Work}
\label{sec:related}
Mesh simplification techniques span from fine-grained vertex-count reduction to coarse shape decomposition \cite{acdpoly, decimation, quadricdecimation}. In this work, we focus on coarse simplification methods involving convex shape decomposition in order to leverage efficient algorithms for robotics and simulation \cite{gjk, convex_motion_planning}.
\subsection{Primitive Fitting}
Researchers have studied how to fit primitives (i.e. cylinders, spheres, boxes) to 3D meshes. Some use learning-based methods to learn important features for identifying shapes \cite{shapefitting,cubes} while others use optimization-based approaches \cite{quadrics}. Primitive fitting has useful applications in robotics, as shown by \cite{foam}, who demonstrated how sphere approximation can be used to accelerate simulation performance. In addition to simulation, simplifying geometry with super-quadric primitives can make pose estimation and grasping more robust \cite{extrusiongrasping, superquadricpose}. 
Fitting geometric primitives often fails to preserve sharp edges and corners. We address this by approximating 3D meshes exclusively with convex hulls, which can adapt to maintain those sharp features.
\subsection{Convex Decomposition} 
Exact Convex Decomposition (ECD) is a method to fit convex hulls over convex regions of the mesh \cite{ECD}. While ECD preserves the original shape exactly, it can produce many parts in trying to preserve all the details in the original shape. Instead of enforcing an exact decomposition, we can relax the requirement by fitting convex hulls approximately within a user-specified error tolerance. 

Approximate Convex Decomposition (ACD) approaches \cite{vhacd, coacd, andrews2024navigation} balance accuracy and performance. Achieving lower approximation error might involve decomposing into more parts whereas decomposing into less parts tolerates more error but may improve simulation or rendering performance. ACD algorithms iteratively select cutting planes on the mesh in order to minimize the approximation error, or ``concavity'' error. V-HACD \cite{vhacd} voxelizes an input mesh and greedily selects cutting planes until either a part count or an error threshold is reached. Instead of greedily selection, Co-ACD \cite{coacd} uses Monte-Carlo Tree Search to look ahead and compute the future approximation error from a sequence of cuts to make more optimal choices. While V-HACD and Co-ACD aim to minimize concavity error using up to a specified number of hulls, Andrews et al. \cite{andrews2024navigation} present an ACD algorithm to preserve “navigable space”, modeled by spheres. Our approach builds on existing ACD algorithms and can internally leverage any suitable method (see Section III.C). In addition to the cutting planes generated by the core ACD algorithm, we also accept user‑provided cuboid regions and analytically derive extra cutting planes from those regions.

Existing ACD methods handle convex decomposition using one uniform error tolerance for the entire mesh, so they cannot honor different tolerances in different regions. This limits their usage in robotics applications, where application-specific knowledge can often be used to improve performance by tuning detail spatially across the mesh. We present an algorithm that overcomes this limitation by accepting a set of selected mesh regions and per-region error parameters as algorithm inputs.

\subsection{Interactive Mesh Simplification}
Automated mesh simplification typically relies on a single threshold to limit over‑ or under‑decomposition, but this global control may be insufficient to tightly constrain approximation error. Semi-automatic methods allow users to have fine-grained control or make design trade-offs while still leveraging automated methods as subroutines. Research on hexahedral remeshing showed that including a user in the loop can be useful to generate high-quality hex meshes \cite{hexmeshuser}. The majority of users in a related study preferred their custom hex meshes compared to automatically generated meshes \cite{hexboxmeshuser}. Similarly, researchers have shown that for mesh-to-spheres conversion, adding a manual tweaking process afterwards overcomes suboptimal solutions from state-of-the-art sphere construction techniques \cite{spheremeshuser}. Building on these insights, we introduce an interactive tool that enables fine-grained control over convex decomposition.

\section{{\ToolName} Tool}
\label{sec:tool}

\begin{figure}
    \centering
  
    \includegraphics[width=\columnwidth]{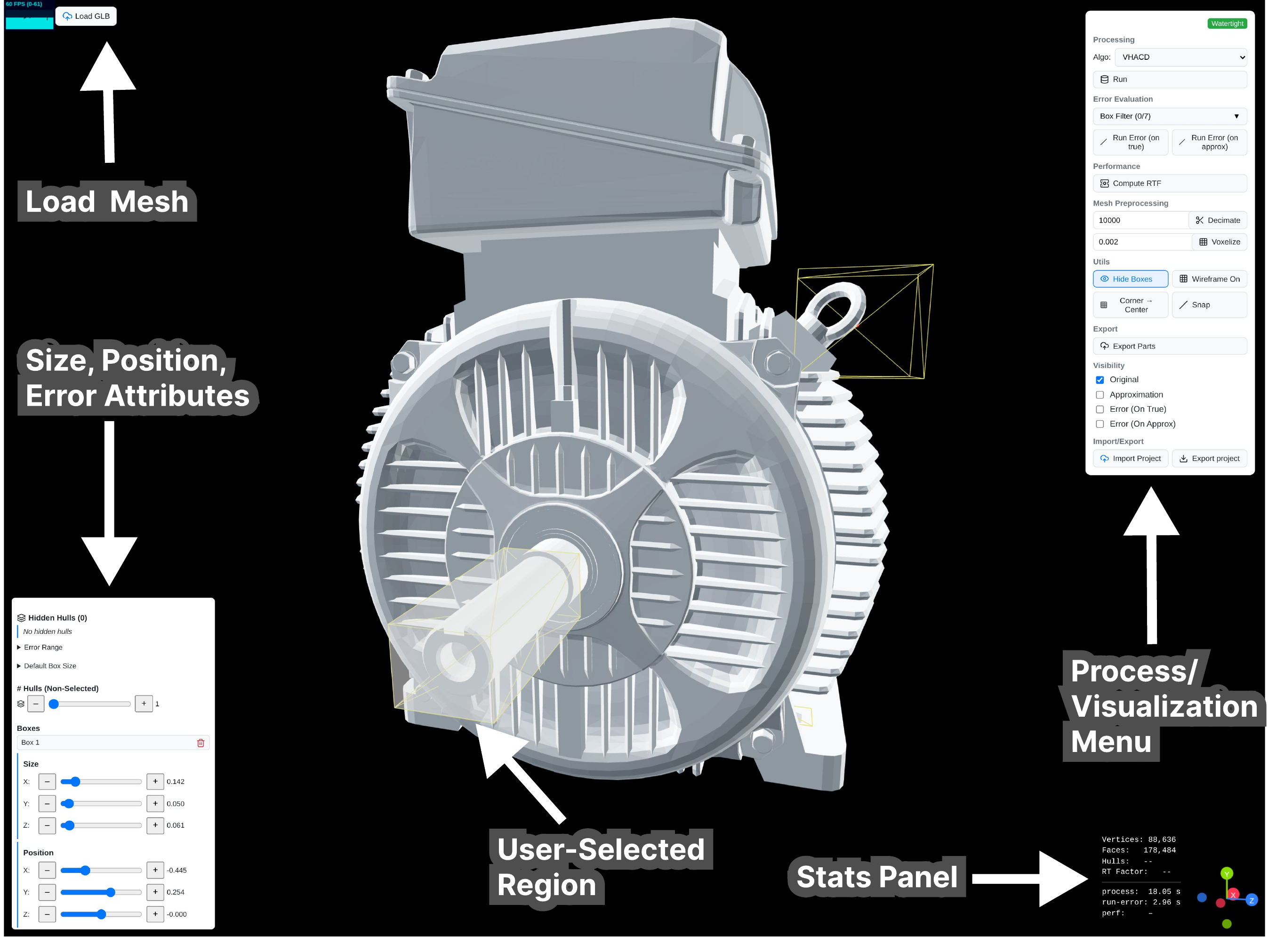}
  
    \caption{
        This figure shows the user interface. The core element is the scene editor, where you can drag and select boxes and pan around the mesh. In the bottom-left panel, you adjust each region’s size, position, and error settings. The top-right panel lets you launch processing services on the mesh and visualize the resulting error metrics.
    } 
    \label{fig:UI}
\end{figure}
The developed interactive mesh simplification tool, called $\ToolName$, enables users to solve a multi-objective optimization problem, trading accuracy and simulation performance.
$\ToolName$ comprises a user interface with visual feedback on the processed meshes and approximation error. The following subsections present the user interface, the optimization problem and the algorithms developed to solve this problem.

\subsection{User Interface}
\subsubsection{Framework}
To enable interactivity, $\ToolName$ is built as a browser-based tool where users can (i) upload a mesh from disk, (ii) simplify the mesh iteratively, (iii) compute comparison metrics between original and simplified mesh(es), and (iv) export the simplified mesh(es) to disk. First, users upload a mesh file through the UI. Then they draw 3D bounding boxes to select regions of the mesh and assign a custom error tolerance to each region. Users then run our interactive convex decomposition algorithm to ensure that each region meets its error-tolerance constraints. Our back-end services provide the convex decomposition algorithm, error computation, and simulation performance metric computation. The final output from the tool is a list of convex meshes that can be imported into other applications. 
 
\subsubsection{Region Selection}
Users can select one or more regions, each containing a part of the mesh, by drawing a 3D bounding box. We constrain region boxes to be axis-aligned in the mesh coordinates, but this is not a key requirement for the algorithms presented below. Drawing boxes in 3D relative to target mesh regions can be tedious. Empart provides a point-and-drag interface to draw a box by specifying a center point and one corner. Optionally, the center point or a face of the box can be snapped to a point on the mesh.

\subsubsection{Visualization}
$\ToolName$ provides several visualizations of the original mesh and approximated convex hulls to assist the user in achieving a satisfactory approximation. Users can hide individual convex hulls to see through the nested layers in the approximation.
A dense approximation error is computed based on the approximated convex hulls and can be visualized as a colored overlay. This error visualization highlights regions of low and high approximation error, indicating potential regions to select for further refinement. Collectively, these visualizations show where the decomposition aligns with or deviates from the user's intent, letting them understand the fit, adjust thresholds, and interactively refine their selected regions.

\subsection{Problem Description}
The interactive convex decomposition algorithm attempts to optimally trade the concavity error and simulation performance with the generated convex part set. The generated approximation must satisfy the specified error tolerance for each selected region. Mathematically, this can be expressed as a constrained non-linear multi-objective optimization problem. Given a watertight triangle mesh \( M=(V,F)\) and a set of selected face regions \(U=\{U_1,\dots,U_R\}\), \emph{interactive convex decomposition} aims to find a partition \( \mathbb{P}=\{P_1,\dots,P_K\}\) of the face set \(F\) that minimizes
\[
  \min_{\mathbb P}\;
  \lambda_{\mathrm{err}}\,\xi(\mathbb P)
  + \lambda_{\mathrm{sim}}\,\tau(\mathbb P)
\]
subject to

\paragraph{Partition}
\[
  P_i\cap P_j = \emptyset \quad (i \neq j),
  \qquad
  \bigcup_{k=1}^{K} P_k = F.
\]
Enforces that the parts are pairwise disjoint (no face overlaps) and exhaustive (all faces are covered).

\paragraph{Region--exclusion}
\[
  \forall\,k,r:\quad P_k \subset U_r \;\lor\; P_k \cap U_r = \emptyset.
\]
Ensures each part lies entirely within a selected region or avoids it, never straddling its boundary.

\medskip

\noindent\textbf{Concavity error term.}
\[
  \xi(\mathbb P)
  = \sum_{r=1}^{R}\bigl(\phi(P_r)-\delta_r\bigr)^2,
\]
measures how far each region’s concavity \(\phi(P_k)\) deviates from the user-specified target concavity \(\delta_k\ge0\), with quadratic penalties for mismatches.

\medskip

\noindent\textbf{Simulation term.}
\(\tau(\mathbb P)\) estimates a simulation performance metric for the approximation, as described in section \ref{subsubsec:sim_perf}.

\medskip

\noindent
The non-negative weights
$(\lambda_{\mathrm{err}}, \,
  \lambda_{\mathrm{sim}})$
shape the trade-off between preserving concavities in the approximate convex decomposition and the resulting simulation performance.

\subsection{Algorithms}

\subsubsection{Interactive Convex Decomposition}
We propose an interactive convex decomposition algorithm (Algorithm~\ref{alg:interactive-alg}) that takes user-specified regions and per-region error tolerances, and produces a decomposition satisfying those requirements. First, a boolean-difference step carves the input mesh into separate partitions. Each partition is then simplified independently using off-the-shelf convex-decomposition routines under its assigned error tolerance (Algorithm~\ref{alg:processbox-alg}). The remaining unselected region is decomposed as well (Algorithm~\ref{alg:decomp-alg}). To prevent any convex part of the unselected region from intruding into the selected region (possibly covering holes or gaps), we apply \texttt{booleanConvexDifference}, which excludes parts overlapping in the selected regions while preserving the convexity of every output piece. Both the mesh-boolean differencing and the convex decomposition of each region are parallelized, yielding sub-linear scaling with respect to the number of selected regions. If any selected region has an error tolerance of zero, the algorithm bypasses convex decomposition and directly returns the partitioned mesh as a special case. We also include a post-processing algorithm \texttt{MergeNeighbors} which merges convex parts together within a specified volume error threshold.
\begin{algorithm}[H]
\caption{\texttt{Interactive\_Convex\_Decomposition}}
\label{alg:interactive-alg}
\begin{algorithmic}[1]
\Require
\Statex \texttt{input\_mesh} $\mathbf{M}$ \hfill (input 3D mesh)
\Statex \texttt{boxes} $\mathbf{B}$ \hfill (list of bounding boxes)
\Statex per-box error tolerance $\varepsilon_i$
\Statex remainder error tolerance $\varepsilon_{\mathrm{rem}}$
\Statex volume-merge tolerance $\tau$

\State $\mathbf{M}_{\mathrm{rem}} \gets \textsc{BooleanDifference}(\mathbf{M},\mathbf{B})$

\State $\mathbf{S}_{\text{convex}} \gets []$ \Comment{List of convex parts}
\State $\mathbf{S}_{\text{mesh}} \gets []$ \Comment{List of meshes}

\ForAll{$b_i \in \mathbf{B}$ with tolerance $\varepsilon_i$}
    \State $(P_b, M_b) \gets \textsc{ProcessBox}(b,\mathbf{M},\varepsilon_i)$
  \State $\mathbf{S}_{\text{convex}}\text{.append}(P_b)$
  \State $\mathbf{S}_{\text{mesh}}\text{.append}(M_b)$
\EndFor

\State $\mathbf{G} \gets \textsc{DecompRemainder}(\mathbf{M}_{\mathrm{rem}}, \mathbf{B}, \varepsilon_{\mathrm{rem}})$

\State $\mathbf{G}_{\text{final}} \gets \textsc{MergeNeighbors}(\mathbf{G} + \mathbf{S}_{\text{convex}}, \tau)$

\State \textbf{return} $\mathbf{G}_{\text{final}},\,\mathbf{S}_{\text{mesh}}$
\end{algorithmic}
\end{algorithm}





  
\begin{algorithm}[H]
\caption{\textsc{ProcessBox}(b, $M$, $\varepsilon$)}
\label{alg:processbox-alg}
\begin{algorithmic}[1]
\Require
\Statex \texttt{box} $b$ \hfill (current bounding box)
\Statex full mesh $M$ \hfill (input 3D mesh)
\Statex error tolerance $\varepsilon$ \hfill (approximation tolerance)

\State $\texttt{select\_convex} \gets []$
\State $\texttt{select\_mesh} \gets []$
\State $M_b \gets \textsc{BooleanIntersect}(M, b)$

\If{$\varepsilon > 0$}
\State $\mathbf{S} \gets \textsc{ConvexDecompose}(M_b, \varepsilon)$
\State $\texttt{select\_convex} \gets \mathbf{S}$
\Else
\State $\texttt{select\_mesh} \gets [M_b]$
\EndIf

\State \Return $\texttt{select\_convex},,\texttt{select\_mesh}$
\end{algorithmic}
\end{algorithm}

\begin{algorithm}[H]
\caption{\textsc{DecompRemainder}($\mathbf{M}_{\mathrm{rem}}, \mathbf{B}, \varepsilon_{\mathrm{rem}})$}
\label{alg:decomp-alg}
\begin{algorithmic}[1]
\Require
\Statex $\mathbf{M}_{\mathrm{rem}}$ \hfill (remaining mesh)
\Statex $\mathbf{B}$ \hfill (list of bounding boxes)
\Statex remainder error tolerance $\varepsilon_{\mathrm{rem}}$

\State $\mathbf{S}_{\text{convex}} \gets \textsc{ConvexDecompose}(\mathbf{M}_{\mathrm{rem}}, \varepsilon_{\mathrm{rem}})$

\ForAll{$b \in \mathbf{B}$}
  \State $\mathbf{S}_{\text{convex}} \gets \textsc{BoolDifferenceConvex}(\mathbf{S}_{\text{convex}}, b)$
\EndFor

\State \Return $\mathbf{S}_{\text{convex}}$
\end{algorithmic}
\end{algorithm}

\subsubsection{Error Visualization}

  
  

\begin{figure}
    \centering
  
    \includegraphics[width=\columnwidth]{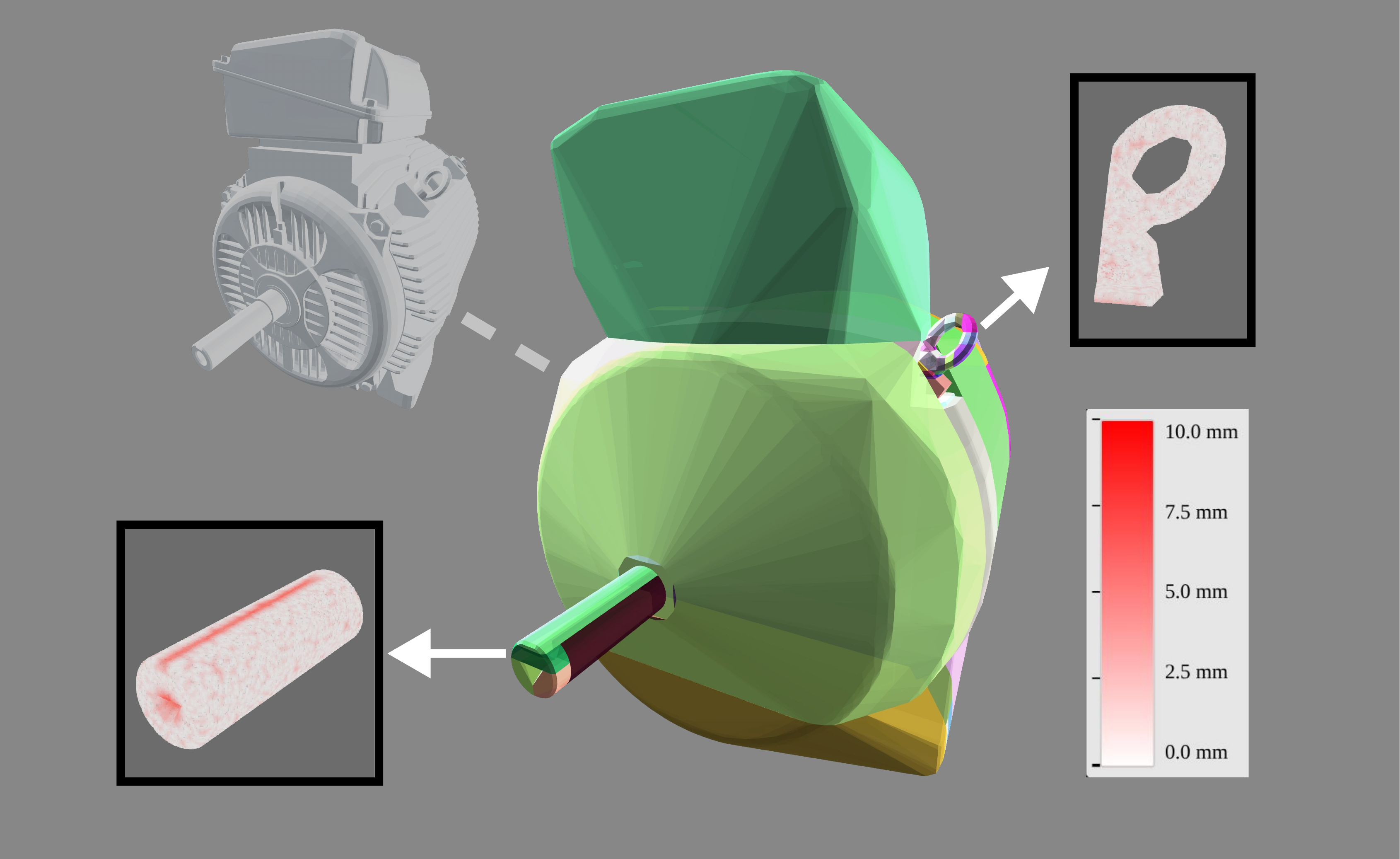}
  
    \caption{
        This graphic shows the approximation in the center and error visualizations of the shaft and lift eye. 
    } 
    \label{fig:error-parts}
\end{figure}

\begin{figure}
    \centering
  
    \includegraphics[width=\columnwidth]{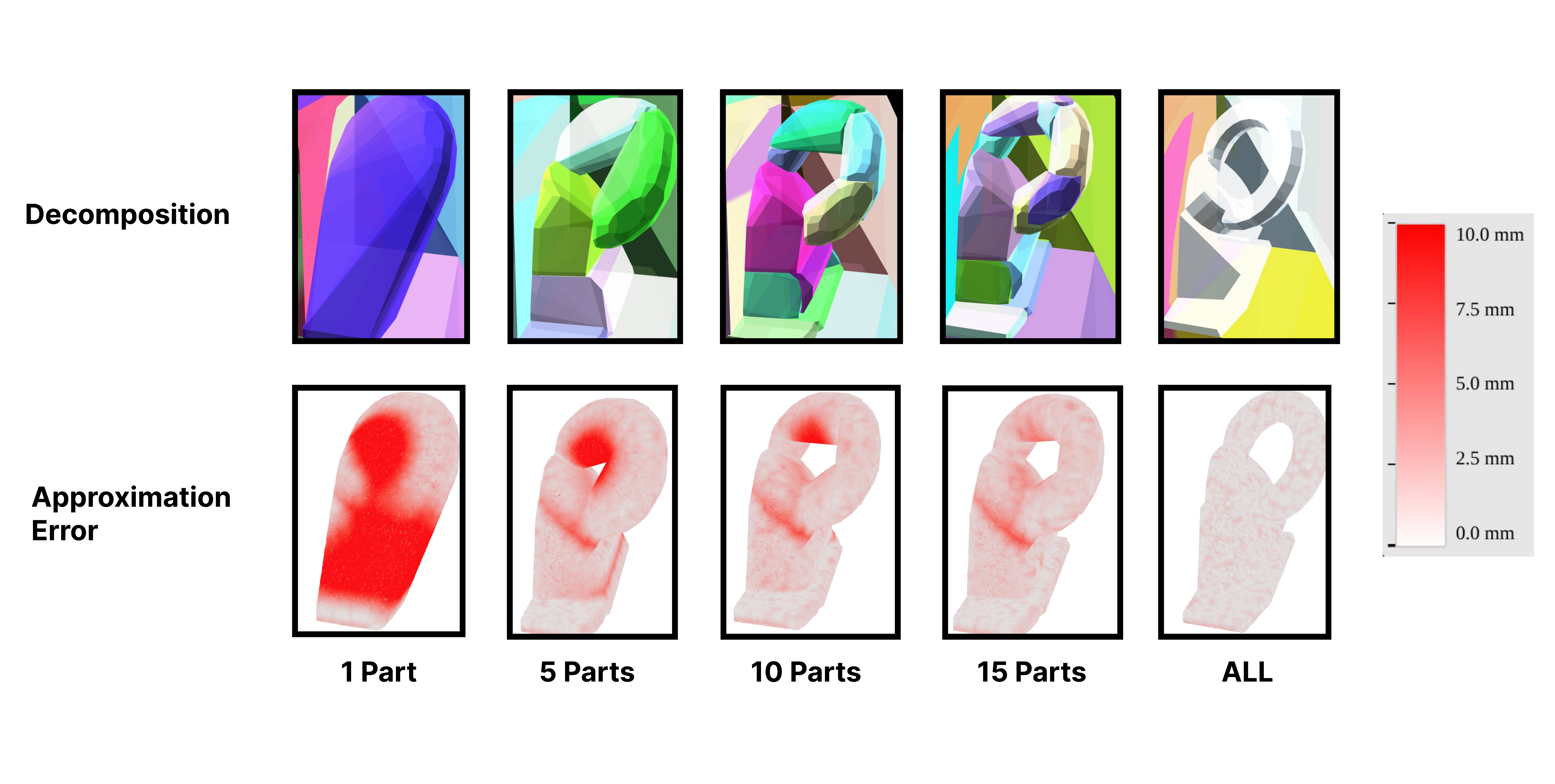}
  
    \caption{
        Top row: approximation of the lifting eye in the original mesh with progressively more convex parts using $\ToolName$. Bottom row: corresponding approximation-error heatmaps over the same region. As the part count increases, error visibly decreases. Selecting the \textit{ALL} option on the UI allows to exactly preserve the original mesh in the select box region.
    } 
    \label{fig:error-seq}
\end{figure}
To evaluate and visualize approximation error, we uniformly sample points on the original mesh and compute the shortest distance from each point to the approximated surface. We compute the approximation-to-original mesh distances as well. These distances are normalized to the [0,1] interval and visualized using a color-map. We apply a boolean addition operation to the convex parts in order to prevent sampling from internal faces. The complete error computation procedure is described in the Appendix (Algorithm~\ref{alg:errorCombined}). As shown in Fig.~\ref{fig:error-seq}, the visualization highlights regions with varying levels of error, red indicating regions with high error (up to 10\,mm in this example) and white indicating regions with low error.

\subsubsection{Simulation Performance}
\label{subsubsec:sim_perf}
\begin{figure}
    \centering
  
    \includegraphics[width=\columnwidth]{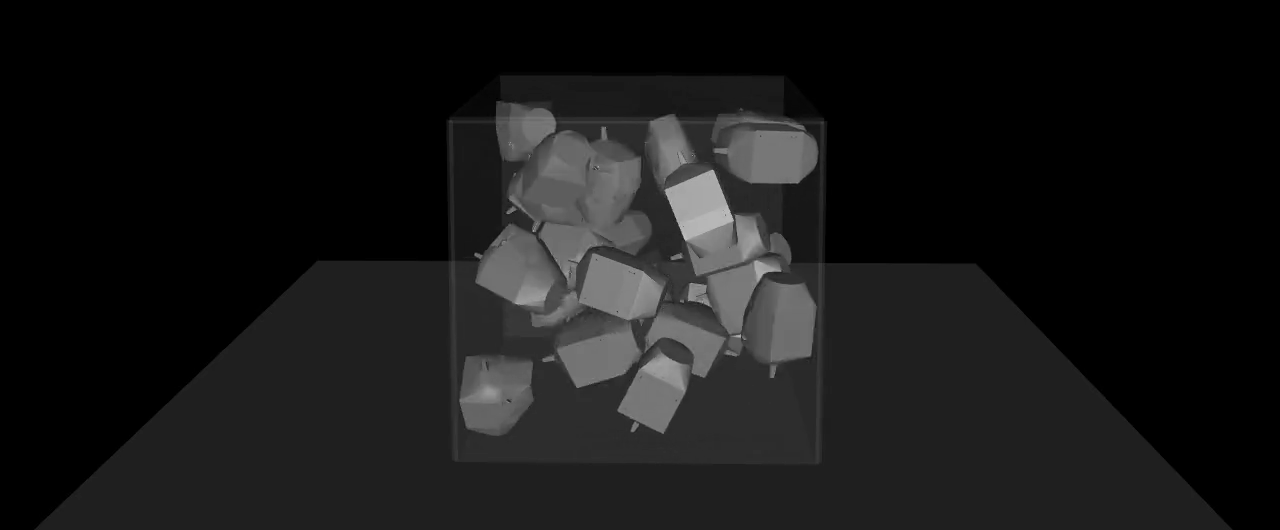}
  
    \caption{
        Each simulation consists of instantiating 25 objects in a uniformly spaced grid inside a closed box such that they are no contacts initially, and then applying randomized forces to the objects over a period of 0.1 seconds. The decomposed parts are combined to construct the collision geometry for each object in the simulation. The returned metrics can be used to evaluate how performant a particular approximation of the original geometry of the object is in simulation in a task-agnostic manner.
    } 
    \label{fig:simperf}
\end{figure}
To assess the simulation efficiency of a decomposition produced by our tool without knowledge of the task, we introduce a task-agnostic performance metric. Using the Mujoco simulator \cite{mujoco}, we apply randomly sampled forces to each object in a 3D array, each represented as a collection of convex primitives (See Fig. ~\ref{fig:simperf}). During the simulation, we log the elapsed simulation time and divide it by the wall-clock time of the entire simulation. This ratio provides a consistent basis for comparing the performance of different decompositions generated by our tool.

\section{Experiments}
\label{sec:experiments}
\begin{figure}
    \centering
    \includegraphics[width=0.5\textwidth]{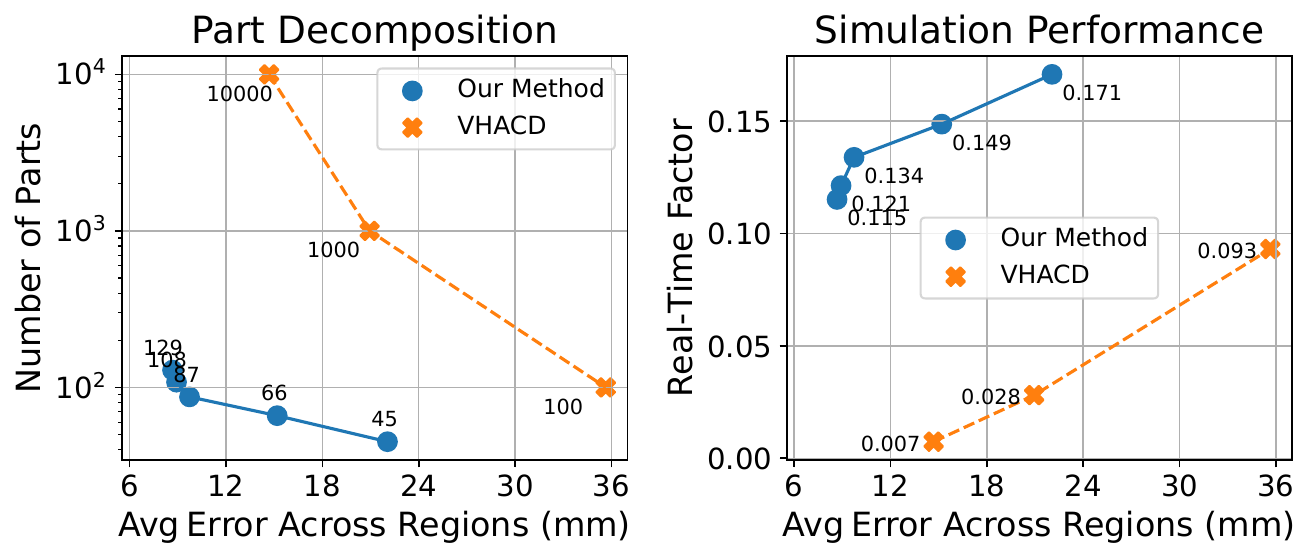}
    \caption{Error is calculated by taking the average across all selected regions' Hausdorff error. We compare our method in blue with V-HACD in orange. We compare the number of parts vs error (left) and simulation real-time factor vs error (right).
}
    \label{fig:res}
\end{figure}
The interactive convex decomposition method presented above is only useful if it improves collision-checking performance over automatic approximate convex decomposition for the same level of preserved detail. We hypothesize:

\begin{description}
  \item[\textbf{H1}] Given an object mesh, a non-empty set of specified regions on the mesh and a maximum tolerable error metric \emph{over the selected regions} (defined below), our method yields \emph{fewer} convex parts than automatic convex decomposition, and a \emph{higher} simulator performance metric.
\end{description}

\textbf{Experimental Design}

\begin{enumerate}
  \item \emph{Region selection} — Partition the input mesh into regions using our tool.
  \item \textbf{Symmetric Hausdorff distance}\\
    For each region we compute two directional Hausdorff errors \cite{hausdorff} before taking their maximum:
    \begin{enumerate}
      \item \textbf{Approx\(\to\)Original:}\\
        Sample points on the \emph{approximated} surface within the region. For each sample, compute its closest-point distance to \emph{anywhere} on the \emph{entire original} mesh. Let
        \[
          d_{a\to o} = \max_{\substack{p \in \text{samples on}\\\text{approx.\ within region}}}
          \min_{q \in \text{original mesh}} \|p - q\|.
        \]
      \item \textbf{Original\(\to\)Approx:}\\
        Sample points on the \emph{original} mesh within the same region. For each sample, compute its closest-point distance to \emph{anywhere} on the \emph{full approximated} surface. Let
        \[
          d_{o\to a} = \max_{\substack{p \in \text{samples on}\\\text{original within region}}}
          \min_{q \in \text{approx.\ mesh}} \|p - q\|.
        \]
      \item The region’s error is
        \[
          \max\bigl(d_{a\to o},\,d_{o\to a}\bigr).
        \]
    \end{enumerate}
    Finally, we average these per-region errors to compute a single overall error metric for the decomposition.
  \item \emph{V-HACD} — Without lack of generalization, we select V-HACD for this experiment for automatic convex decomposition. We use maximum convex hull counts of 100, 1000, and 10000 for V-HACD and compute the above error for each approximation.
  \item \emph{Our method} — We vary each region’s part budget from 1 to 13 in increments of 3, and plot the resulting trade-off curves for (i) total part count versus error and (ii) simulation real-time factor versus error.
\end{enumerate}

To evaluate our approach, we applied it to a motor mesh and annotated seven region selections that isolate key features, namely the shaft, four mounting holes and two lifting eyes. The resulting metrics are shown in Fig.~\ref{fig:res}. Reaching an average Hausdorff distance of 13 mm requires V-HACD to generate 10000 parts compared to 100 parts in our method. At this error threshold, simulation performance is also 20x better for our method compared to V-HACD, which supports hypothesis \textbf{H1}. Note: The real-time factor for VHACD with 100 parts (0.093) is lower than for our method with 129 parts (0.115). This suggests that efficiency depends not only on part count but also on how hulls participate in narrow-phase checks; VHACD distributes contact across many hulls, while our method concentrates it into fewer areas.

\section{Application}
\label{sec:application}
To investigate the applicability of Empart in robotics simulations, we defined an example robotic pick-and-place scenario in Gazebo (with Bullet multibody physics engine) \cite{Gazebo}.
The scenario consists of picking motors from a stationary table with a robot arm and placing them on another stationary table.
Each motor is placed on a mounting plate with four protruding studs that are inserted into four mounting holes at the bottom of the motor.
The scenario is simulated once with collision geometry for the motor generated from Empart, where a Hausdorff error of 5mm was accepted for the mesh region adjacent to and including each mounting hole (regions are selected in Empart as explained in Section IV).
Subsequently, the simulation is repeated with collision geometry for the motor generated using uniform V-HACD-decomposition with a maximum hull count of 10000, voxel resolution of 5e5 and maximum acceptable volume error of 0.001\% to preserve the mounting holes accurately.
The aim of this comparison is to investigate the relative impact of refining collision geometry using Empart on simulation performance for a representative robotics application.
That is, does the task-agnostic simulation performance improvement observed in Section IV translate into faster application simulation?

\begin{figure*}[!t]
    \centering
    \includegraphics[width=\textwidth]{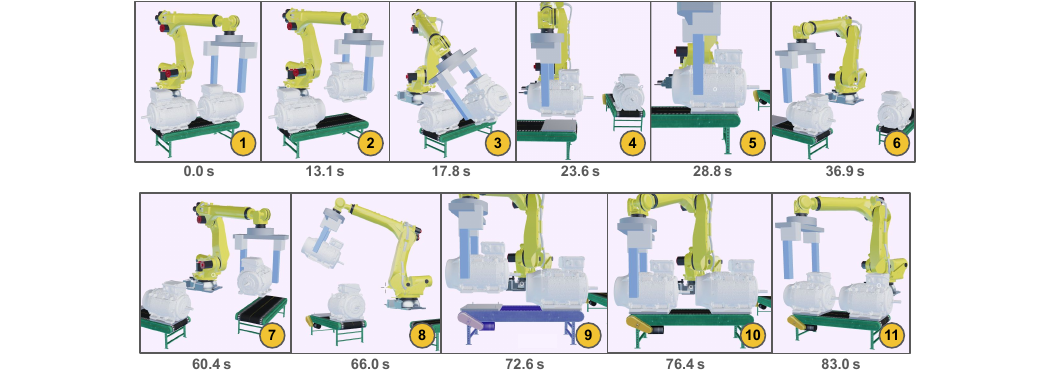}
    \caption{Pick-and-place application simulation showing key time-points (in simulation time). (1) start, (2) grasped and picked up motor 1, (3) moving motor 1 to placement pose (motors are in close proximity), (4) reached placement pose, (5) placed motor 1 on mounting plate with studs, (6) released motor 1, (7) grasped and picked up motor 2, (8) moving motor 2 to placement pose, (9) reached placement pose, (10) placed motor 2 on mounting plate with studs, and (11) released motor 2, end of motion. Two simulations are run for this application - first with motor collision meshes generated using Empart and next with meshes generated using V-HACD. }
    \label{fig:application}
\end{figure*}

The pick-and-place application involves a stationary robot arm (Fanuc R-2000iC/210L) grasping one motor at a time and placing it on a mounting plate.
This process is repeated for two motors.
The overall application program is defined using a behavior tree, composed of the following nodes - \textit{move robot with collision checking, move robot linear, grasp} and \textit{release} \cite{behaviortree}.
Robot motions are generated using a motion planner to ensure collision-free paths and interpolated using quintic polynomials to generate smooth acceleration-limited trajectories.
A 2-fingered pinch gripper with prismatic joints is used to grasp the motor.
Robot and gripper joint motions are executed in Gazebo using a PID joint-position controller superimposed with gravity compensation torque at each joint.
To reduce the number of pairwise contact interactions that affect the simulation, the gripper fingers are constrained through a position equality constraint to be affixed to the motor after initial contact. This effectively fixes the gripper to the motor for the duration of the grasp.
The position equality constraints are removed once the motor is released from the gripper.
Figure \ref{fig:application} shows a series of simulation time-points from Gazebo, from application start (T1) to finish (T11).
All simulations were run on an \textit{N1-standard-8} VM on Google Cloud Compute Engine \cite{gcloud_compute_engine}.

Table \ref{tab:application} compares the application simulation performance between Empart and V-HACD collision meshes in terms of the overall real-time factor as well as the "current" real-time factor measured at different time-points in the application.
The "current" real-time factor (RTF) is a time-varying measure of simulation performance, computed as the ratio of simulation-clock elapsed time to wall-clock elapsed time over a rolling window of 1\,s (on wall clock).
Gazebo was configured to run with a time step of 2\,ms and at a target current RTF of 1.0, to visualize robot motions as close to real-time as possible.
As a result, the current RTF measured during the simulations never exceeded 1.0. 
Overall, the application could be simulated accurately with Empart collision meshes within 106\,s (at a slowdown of 0.78x real-time), where as the same task required 377\,s to simulate using V-HACD collision meshes (at a slowdown of 0.22x real-time).
The minimum current RTF was observed in both simulations when either motor was placed on the mounting plate --- 0.34 for Empart collision meshes, compared to 0.07 for V-HACD collision meshes.
Simulation performs poorly at this time-point due to the large number of collisions that must be resolved between the mounting holes and the studs on the mounting plates for the motor.
At each observed time-point, current RTF was higher for Empart collision meshes compared to V-HACD collision meshes.
The difference between current RTF for the two simulations was lowest at time-points T1 and T7, where simulator optimizations such as island deactivation and broad-phase collision checking reduced the computational burden of collision checking.

In summary, for this representative robotic pick-and-place application, refining collision meshes for the motor object using Empart reduced simulation time by 69\%.
Not only did the overall simulation run faster with Empart collision meshes, but also the simulation ran faster at each time-point compared to using uniformly-decomposed convex collision meshes.
Particularly at time-point T10 where a large number of collisions had to be resolved in each simulation, using Empart collision meshes sped up simulation by 4.9 times.
These results emphasize the potential for significant improvement in simulation performance for robotics applications by employing Empart-generated collision meshes.
At the same time, this experiment demonstrates the non-linear relationship between the number of convex hulls used for collision checking and the current RTF due to various optimizations in modern robotics simulators.
Consequently, Empart is most suitable for refining collision meshes in an iterative fashion, with performance feedback from application simulation.

\begin{table*}[!t]
    \centering
    \caption{Overall and Current Real-time factor (RTF) comparison for simulated application}
    \label{tab:application}
    \begin{tabular}{|l|c| *{11}{c} |}
        \toprule
        \multirow{2}{*}{Motor collision meshes source} & \multirow{2}{*}{Overall RTF} & \multicolumn{11}{c|}{Current RTF at the time-points indicated in Fig. \ref{fig:application}. (Min values are italicized)} \\ 
        \cmidrule(l){3-13} 
        & & T1 & T2 & T3 & T4 & T5 & T6 & T7 & T8 & T9 & T10 & T11 \\
        \midrule
        Empart & 0.78 & 0.98 & 0.98 & 0.85 & 0.97 & \textit{0.34} & 0.71 & 0.98 & 0.42 & 0.98 & \textit{0.34} & 0.72 \\
        V-HACD & 0.22 & 0.94 & 0.71 & 0.40 & 0.70 & \textit{0.07} & 0.20 & 0.83 & 0.12 & 0.65 & \textit{0.07} & 0.07 \\
        \bottomrule
    \end{tabular}
\end{table*}

\section{Conclusion}

\label{sec:conclusion}
We've demonstrated that our interactive convex decomposition tool, $\ToolName$, addresses a critical limitation of existing mesh simplification methods: the inability to apply task-specific spatial detail.
By allowing users to define and simplify specific regions of a mesh with custom error tolerances and providing immediate feedback through intuitive visualizations and performance metrics, $\ToolName$ produces significantly more efficient and task-appropriate collision geometry.
Our evaluation showed that this fine-grained control leads to a dramatic reduction in the number of convex parts, resulting in substantial simulation speedups.
For a representative robotics task, our method improved simulation performance by nearly 70\% compared to a uniform decomposition.
We are thrilled to release $\ToolName$ as an open-source project \footnote{Source code is available on Github at https://github.com/intrinsic-opensource/empart.git}, and we invite the community to explore its capabilities and contribute to its development.

\textbf{Future work.} We plan to expand $\ToolName$ to support decompositions into analytic shape primitives. We also plan to explore usability improvements for region selection, potentially through semantic segmentation of the original mesh. The current prototype already measures the simulation speed‑ups gained from a single simplified mesh; the same approach can be extended to additional metrics such as collision-detection for motion planning. By presenting performance data alongside approximation error, $\ToolName$ will help users optimize their geometry more quickly and boost their application performance.

\addtolength{\textheight}{-12cm}   







{\footnotesize
\bibliographystyle{IEEEtran}
\bibliography{references}

\begin{thebibliography}{10}
\providecommand{\url}[1]{#1}
\csname url@samestyle\endcsname
\providecommand{\newblock}{\relax}
\providecommand{\bibinfo}[2]{#2}
\providecommand{\BIBentrySTDinterwordspacing}{\spaceskip=0pt\relax}
\providecommand{\BIBentryALTinterwordstretchfactor}{4}
\providecommand{\BIBentryALTinterwordspacing}{\spaceskip=\fontdimen2\font plus
\BIBentryALTinterwordstretchfactor\fontdimen3\font minus \fontdimen4\font\relax}
\providecommand{\BIBforeignlanguage}[2]{{%
\expandafter\ifx\csname l@#1\endcsname\relax
\typeout{** WARNING: IEEEtran.bst: No hyphenation pattern has been}%
\typeout{** loaded for the language `#1'. Using the pattern for}%
\typeout{** the default language instead.}%
\else
\language=\csname l@#1\endcsname
\fi
#2}}
\providecommand{\BIBdecl}{\relax}
\BIBdecl

\bibitem{acdpoly}
\BIBentryALTinterwordspacing
J.-M. Lien and N.~M. Amato, ``Approximate convex decomposition of polygons,'' in \emph{Proceedings of the Twentieth Annual Symposium on Computational Geometry}, ser. SCG '04.\hskip 1em plus 0.5em minus 0.4em\relax New York, NY, USA: Association for Computing Machinery, 2004, p. 17–26. [Online]. Available: \url{https://doi.org/10.1145/997817.997823}
\BIBentrySTDinterwordspacing

\bibitem{decimation}
\BIBentryALTinterwordspacing
W.~J. Schroeder, J.~A. Zarge, and W.~E. Lorensen, ``Decimation of triangle meshes,'' in \emph{Proceedings of the 19th Annual Conference on Computer Graphics and Interactive Techniques}, ser. SIGGRAPH '92.\hskip 1em plus 0.5em minus 0.4em\relax New York, NY, USA: Association for Computing Machinery, 1992, p. 65–70. [Online]. Available: \url{https://doi.org/10.1145/133994.134010}
\BIBentrySTDinterwordspacing

\bibitem{quadricdecimation}
\BIBentryALTinterwordspacing
M.~Garland and P.~S. Heckbert, \emph{Surface Simplification Using Quadric Error Metrics}, 1st~ed.\hskip 1em plus 0.5em minus 0.4em\relax New York, NY, USA: Association for Computing Machinery, 2023. [Online]. Available: \url{https://doi.org/10.1145/3596711.3596727}
\BIBentrySTDinterwordspacing

\bibitem{gjk}
E.~Gilbert, D.~Johnson, and S.~Keerthi, ``A fast procedure for computing the distance between complex objects in three-dimensional space,'' \emph{IEEE Journal on Robotics and Automation}, vol.~4, no.~2, pp. 193--203, 1988.

\bibitem{convex_motion_planning}
J.~Schulman, Y.~Duan, J.~Ho, A.~Lee, I.~Awwal, H.~Bradlow, J.~Pan, S.~Patil, K.~Goldberg, and P.~Abbeel, ``Motion planning with sequential convex optimization and convex collision checking,'' \emph{The International Journal of Robotics Research}, vol.~33, no.~9, pp. 1251--1270, 2014.

\bibitem{shapefitting}
\BIBentryALTinterwordspacing
Y.~Li, S.~Liu, X.~Yang, J.~Guo, J.~Guo, and Y.~Guo, ``Surface and edge detection for primitive fitting of point clouds,'' in \emph{ACM SIGGRAPH 2023 Conference Proceedings}, ser. SIGGRAPH '23.\hskip 1em plus 0.5em minus 0.4em\relax New York, NY, USA: Association for Computing Machinery, 2023. [Online]. Available: \url{https://doi.org/10.1145/3588432.3591522}
\BIBentrySTDinterwordspacing

\bibitem{cubes}
\BIBentryALTinterwordspacing
C.-Y. Sun, Q.-F. Zou, X.~Tong, and Y.~Liu, ``Learning adaptive hierarchical cuboid abstractions of 3d shape collections,'' \emph{ACM Trans. Graph.}, vol.~38, no.~6, 2019. [Online]. Available: \url{https://doi.org/10.1145/3355089.3356529}
\BIBentrySTDinterwordspacing

\bibitem{quadrics}
T.~Birdal, B.~Busam, N.~Navab, S.~Ilic, and P.~Sturm, ``A minimalist approach to type-agnostic detection of quadrics in point clouds,'' in \emph{Proceedings of the IEEE Conference on Computer Vision and Pattern Recognition}, 2018, pp. 3530--3540.

\bibitem{foam}
\BIBentryALTinterwordspacing
S.~Coumar, G.~Chang, N.~Kodkani, and Z.~Kingston, ``Foam: A tool for spherical approximation of robot geometry,'' 2025. [Online]. Available: \url{https://arxiv.org/abs/2503.13704}
\BIBentrySTDinterwordspacing

\bibitem{extrusiongrasping}
A.~Huaman~Quispe, B.~Milville, M.~Gutierrez, C.~Erdogan, M.~Stilman, H.~Christensen, and H.~Ben~Amor, ``Exploiting symmetries and extrusions for grasping household objects,'' in \emph{Proceedings - IEEE International Conference on Robotics and Automation}, vol. 2015, 05 2015.

\bibitem{superquadricpose}
\BIBentryALTinterwordspacing
X.~Tu and K.~Desingh, ``Superq-grasp: Superquadrics-based grasp pose estimation on larger objects for mobile-manipulation,'' 2025. [Online]. Available: \url{https://arxiv.org/abs/2411.04386}
\BIBentrySTDinterwordspacing

\bibitem{ECD}
\BIBentryALTinterwordspacing
J.~M. Keil, ``Decomposing a polygon into simpler components,'' \emph{SIAM Journal on Computing}, vol.~14, no.~4, pp. 799--817, 1985. [Online]. Available: \url{https://doi.org/10.1137/0214056}
\BIBentrySTDinterwordspacing

\bibitem{vhacd}
K.~Mamou, E.~Lengyel, and A.~Peters, ``Volumetric hierarchical approximate convex decomposition,'' \emph{Game engine gems}, vol.~3, pp. 141--158, 2016.

\bibitem{coacd}
X.~Wei, M.~Liu, Z.~Ling, and H.~Su, ``Approximate convex decomposition for 3d meshes with collision-aware concavity and tree search,'' \emph{ACM Transactions on Graphics (TOG)}, vol.~41, no.~4, pp. 1--18, 2022.

\bibitem{andrews2024navigation}
J.~Andrews, ``Navigation-driven approximate convex decomposition,'' in \emph{ACM SIGGRAPH 2024 Conference Papers}, 2024, pp. 1--9.

\bibitem{hexmeshuser}
\BIBentryALTinterwordspacing
L.~Li, P.~Zhang, D.~Smirnov, S.~M. Abulnaga, and J.~Solomon, ``Interactive all-hex meshing via cuboid decomposition,'' \emph{ACM Trans. Graph.}, vol.~40, no.~6, Dec. 2021. [Online]. Available: \url{https://doi.org/10.1145/3478513.3480568}
\BIBentrySTDinterwordspacing

\bibitem{hexboxmeshuser}
F.~Zoccheddu, E.~Gobbetti, M.~Livesu, N.~Pietroni, and G.~Cherchi, ``Hexbox: interactive box modeling of hexahedral meshes,'' in \emph{Computer Graphics Forum}, vol.~42, no.~5.\hskip 1em plus 0.5em minus 0.4em\relax Wiley Online Library, 2023, p. e14899.

\bibitem{spheremeshuser}
\BIBentryALTinterwordspacing
D.~Paolillo and M.~Tarini, ``Automatic and user-assisted sphere-mesh construction,'' \emph{IEEE Comput. Graph. Appl.}, vol.~44, no.~6, p. 105–117, Nov. 2024. [Online]. Available: \url{https://doi.org/10.1109/MCG.2024.3426656}
\BIBentrySTDinterwordspacing

\bibitem{mujoco}
E.~Todorov, T.~Erez, and Y.~Tassa, ``Mujoco: A physics engine for model-based control,'' in \emph{2012 IEEE/RSJ international conference on intelligent robots and systems}.\hskip 1em plus 0.5em minus 0.4em\relax IEEE, 2012, pp. 5026--5033.

\bibitem{hausdorff}
D.~Burago, Y.~Burago, and S.~Ivanov, \emph{A Course in Metric Geometry}, ser. Graduate Studies in Mathematics.\hskip 1em plus 0.5em minus 0.4em\relax American Mathematical Society, 2001, vol.~33.

\bibitem{Gazebo}
\BIBentryALTinterwordspacing
{Open Robotics}, ``{Gazebo: A Robotics Simulator},'' Gazebo Documentation, 2025, accessed on August 4, 2025. [Online]. Available: \url{https://gazebosim.org}
\BIBentrySTDinterwordspacing

\bibitem{behaviortree}
M.~Colledanchise and P.~{\"O}gren, \emph{Behavior trees in robotics and AI: An introduction}.\hskip 1em plus 0.5em minus 0.4em\relax CRC Press, 2018.

\bibitem{gcloud_compute_engine}
\BIBentryALTinterwordspacing
{Google Cloud}, ``{Compute Engine Machine Families},'' Google Cloud Documentation, 2025, accessed on August 4, 2025. [Online]. Available: \url{https://cloud.google.com/compute/docs/machine-resource}
\BIBentrySTDinterwordspacing

\end{thebibliography}
}

\newpage
\appendix
\begin{algorithm}[H]

    \caption{\textsc{BoolDifferenceConvex}$(\texttt{P, box})$}
    \begin{algorithmic}
    \Require
\Statex  $P$ \hfill (list of convex parts)
\Statex $\texttt{box}$ \hfill (single box)

        \State $\mathcal{G}_{\text{final}} \gets \emptyset$

\ForAll{$p \in \mathbb{P}$}
    \State $w \gets [\mathbb{M_{\mathrm{rem}}}]$
        \ForAll{plane $(n, d)$ in $\textsc{BoundingBoxPlanes}(b)$}
            \State $w_{\text{next}} \gets \emptyset$
            \ForAll{$\text{piece} \in w$}
                \If{\textsc{BoolDiff(piece, cube)} is $\emptyset$}
                
                \State $w_{\text{next}} \gets w_{\text{next}} \cup \{\text{piece}\}$ 
                    \State \textbf{continue}
                \EndIf
               \State $w_{\mathrm{next}} \gets w_{\mathrm{next}} \;\cup\;\text{piece.split\_by\_plane}(n, d)$
            \EndFor
            \State $w \gets w_{\text{next}}$
        \State $w \gets \{ f \in w \mid \neg \textsc{FullyInside}(f, cube) \}$ \Comment{remove parts fully inside cube}
    \EndFor
    \State $w \gets \textsc{MergeNeighbors}(w, \tau = 0)$
    \State $\mathbb{G}_{\text{final}} \gets \mathbb{G}_{\text{final}} \cup w$
\EndFor
\State \Return $\mathbb{G}_{\text{final}}$

    \end{algorithmic}
\end{algorithm}

\begin{algorithm}[H]
\caption{\texttt{Error\_Samples}}
\label{alg:errorCombined}
\begin{algorithmic}[1]
\Require
\Statex Original surface mesh $\mathbb{T}$
\Statex Approximating meshes $\{\mathbb{A}_1,\dots,\mathbb{A}_k\}$
\Statex Sample count $N$, colormap $\mathbb{C}$, normalization $(\alpha,\beta)$
\Statex (Optional) filter boxes $\mathbb{R}$
\Statex Boolean flag $\mathsf{onApprox}$

\State $\mathbb{A} \gets \textsc{BooleanUnion}(\{\mathbb{A}_i\})$
\If{$\mathsf{onApprox}$}
    \State $P_{\mathbb{T}} \gets \textsc{SampleSurface}(\mathbb{T}, N)$
    \State $\mathbb{I} \gets \textsc{BuildKDTree}(P_{\mathbb{T}})$
    \State $Q \gets \textsc{SampleSurface}(\mathbb{A}, N)$
\Else
    \State $P_{\mathbb{A}} \gets \textsc{SampleSurface}(\mathbb{A}, N)$
    \State $\mathbb{I} \gets \textsc{BuildKDTree}(P_{\mathbb{A}})$
    \State $Q \gets \textsc{SampleSurface}(\mathbb{T}, N)$
\EndIf
\ForAll{point $q \in Q$}
    \State $d_q \gets \textsc{NearestDistance}(q,\mathbb{I})$
    \State $e_q \gets \textsc{NormalizeClamp}(d_q,\alpha,\beta)$
    \State $\text{color}(q) \gets \mathbb{C}(e_q)$
\EndFor
\If{$\neg \mathsf{onApprox}$ \textbf{and} $\mathbb{R} \neq \emptyset$}
    \State $Q \gets \textsc{CutRegions}(Q,\mathbb{R})$
\EndIf
\State \Return $(Q,\text{color})$ \Comment{sample set with per-point colour}
\end{algorithmic}
\end{algorithm}

\end{document}